\def\year{2019}\relax
\documentclass[letterpaper]{article} 
\usepackage{aaai19_mod}  
\usepackage{times}  
\usepackage{helvet}  
\usepackage{courier}  
\usepackage{url}  
\usepackage{graphicx}  
\begin{document}
%
\title{Classification of X-Ray Protein Crystallization \\ Using Deep Convolutional Neural Networks with a Finder Module}
\author{Yusei Miura\textsuperscript{1,2}, Tetsuya Sakurai\textsuperscript{1}, Claus Aranha\textsuperscript{1,2}, Toshiya Senda\textsuperscript{3}, Ryuichi Kato\textsuperscript{3}, Yusuke Yamada\textsuperscript{3}\\
\textsuperscript{1}{University of Tsukuba, Department of Computer Sciences, Japan}\\
\textsuperscript{2}{High Energy Accelerator Research Organization, Japan}\\
\textsuperscript{3}{contact: miura@mma.cs.tsukuba.ac.jp, caranha@cs.tsukuba.ac.jp}\\
}

\maketitle
\begin{abstract}
Recently, deep convolutional neural networks have shown good results for image recognition.
In this paper, we use convolutional neural networks with a finder module, which discovers the important region for recognition and extracts that region.
We propose applying our method to the recognition of protein crystals for X-ray structural analysis.
In this analysis, it is necessary to recognize states of protein crystallization from a large number of images.
There are several methods that realize protein crystallization recognition by using convolutional neural networks.
In each method, large-scale data sets are required to recognize with high accuracy.
In our data set, the number of images is not good enough for training CNN.
The amount of data for CNN is a serious issue in various fields.
Our method realizes high accuracy recognition with few images by discovering the region where the crystallization drop exists.
We compared our crystallization image recognition method with a high precision method using Inception-V3.
We demonstrate that our method is effective for crystallization images using several experiments.
Our method gained the AUC value that is about $5\%$ higher than the compared method.
\end{abstract}

\section{Introduction}
In recent years, deep learning has demonstrated success in several fields.
Deep convolutional neural networks (CNNs) was successful in computer vision recognition tasks \cite{He2015ImageNet,Ioffe2015Batch}.
As CNN began showing achievements, it has been applied to products owned by companies and research institutes.

\begin{figure}[htbp]
  \centering
  \includegraphics[width=0.8\linewidth]{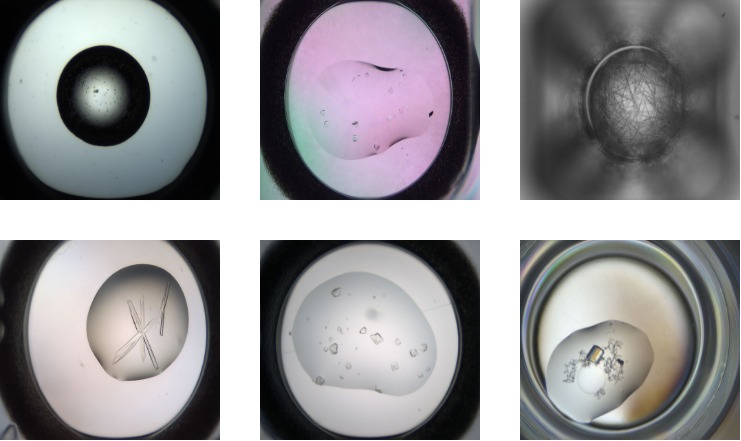}
  \caption{Examples of Crystal images from different facilities.
  From top to bottom, left to right: the Collaborative Crystallization Center, GlaxoSmithKline, Hauptman Woodward Medical Research Institute, Merck \& Co., Bristol-Myers Squibb, and the High Energy Accelerator Research Organization.
  Most crystalliztion image analysis methods focus on only one particular data set.}
  \label{fig:crystalimages}
\end{figure}

X-ray protein crystallograpy is one of CNN application examples \cite{Bruno2018Google}.
X-ray protein crystallography is a powerful technique for determining the three-dimensional structure of protein molecule, which is important to understand the protein's function, based on X-ray diffraction from the crystallized protein.
To obtain a protein crystal, it is necessary to find a crystallization condition for each protein.
However, this process is not straightforward and hundreds or thousands of crystallization conditions are typically examined to find an optimum one.
There are many high throughput crystallization facilities around the world, where making crystallization drops with various crystallization condition and taking pictures of these drops periodically are performed automatically.
Even in such kind of facilities, however, pictures of crystallization drops are evaluated manually.
Several attempts have been made so far to automate the evaluation of crystallization drops by an image recognition method \cite{Pusey2017Protein}.
Given that each facility uses a different imager, crystallization tray, lighting system, and so on (Figure\ref{fig:crystalimages}), and because each of these parameters severely affect image recognition, it is difficult to recognize images with high accuracy and versatility.
CNN can recognize the state of crystallization regardless of such complicated conditions.

The performance demonstrated by CNN is comparable to human discrimination performance, but a large number of images are required.
It is possible to acquire a large number of images in the process of crystallization, however, the labeling of each image must be performed by a person.
Since the evaluation criteria are different for each method, it is difficult to create a data set in which correspondence between images and labels is guaranteed.
The larger the data set is used, the more likely it is for discrepancy between image and label will occur.

In performing image recognition, the whole of the image does not necessarily contribute to improvement in accuracy of recognition.
There are cases in which some areas of the image are particularly important for evaluation.
For example, only the "drop" area is necessary for automatic evaluation of a crystallization.
We assume that when a CNN is applied to the whole image, it has to learn how to process the non-important areas as well, and this causes a burden to the learning process.
The result is that a larger data set is necessary for achieving high accuracy.
Therefore we think that by applying a process that can try to discover the important areas beforehand, the CNN can achieve higher precision without learning with a large number of images.

In this paper, we propose applying a method of object classification with a finder module and CNN for crysallization images.
The Finder module discovers the target in the image using U-Net and extracts the area of interest.
Discovering the target prevents the parts irrelevant to the evaluation from affecting the recognition.
CNN can get important features efficiently after using a finder module.
We attempted automatic evaluation of crystallization images possessed by the High Energy Accelerator Research Organization (KEK).
By applying the proposed method to the crystallization images, it was successfully demonstrated that the proposed method produced better results than the method using only CNN on the full image.

\section{Related Work}
CNNs \cite{Fukushima1980Neo,LeCun1989BP} are neural networks mainly used in image recognition.
For example, Krizhevsky et al. \shortcite{Krizhevsky2012ImageNet} won the ImageNET Large Scale Visual Recognition Competition (ILSVRC) using CNN.
Whereas a neural network for images loses the spatial information, the CNN has special characteristics that help preserving that information.
One is the introduction of local receptive fields by the convolution layer (i.e., the convolution of filters for an input to acquire features).
In the process of convolution, weight sharing is also performed to suppress parameters.
The other is to apply down sampling by the pooling layer.
The pooling layer acquires an invariance on the change of object position in the image.

CNN is used not only for image classification (labeling images), but also for image segmentation (finding parts of an image) and object detection (finding objects of an image).
Regarding CNN application to image segmentation, Fully Convolutional Network \cite{Long2015FCN} (FCN) which is composed only of convoliton layers is often used.
On the other hand, R-CNN \cite{Girshick2014Rich} and SSD \cite{Liu2016SSD} is used for object detection.
There is a recognition method that combines some models based on CNN.
For example, in the task of evaluating skin cancer, Chang \shortcite{Chang2017Skin} proposed using FCN for detection of the skin cancer region and CNN for classification.

The method using CNN is also used in the automatic recognition of crystallization images.
Yann et al. \shortcite{Yann2016CrystalNet} proposed CNN named CrystalNet which is composed of 4 convolution layers, 3 pooling layers, and 2 fully connected layers.
Bruno et al. \shortcite{Bruno2018Google} applied a model called Inception-V3 to the huge crystallization data set MARCO (https://marco.ccr.buffalo.edu).
Ghafurian et al. \shortcite{Ghafurian2018ClassificationOP} investigated how different performances would show when various CNNs were applied to their data set.
The criteria of the crystallization state also differ for each facilities and for each recognition method.
Yann et al. attempted classification based on two values, whether or not there are crystals, and finely divided into 10 states (Clear, Precipitate, Crystal, Phase, Precipitate \& Crystal, Precipitate \& skin, Phase \& crystal, Phase \& Precipitate, Skin, Junk).
Bruno et al. divided into 4 states with Clear, Crystals, Precipitate, Others.
Ghafurian et al. divided into 10 states, but the particular states are different from Yann et al.

In the protein crystallization field, CNNs show better result than other crystallization image recognition methods \cite{Zuk1991Protein,Bern2004Curve,Cumbaa2010Protein,Dinc2014Random}.
However, when CNN learns the training data, CNN recognizes the whole of the image including unnecessary parts.
We think that sending the unnecessary part of the images cause the CNN to require a larger number of images for training.
In recognizing the state of crystallization, we think that all the necessary information is contained in the crystallization "drop".
Therefore, we think that it is possible to increase CNN performance by finding the drop in the image first and only giving the drop to the CNN.

In this study, we use FCN for discovering the crystallization drop, and apply CNN to recognition.
We think that FCN is suitable for catching the shape of a drop.
We call the entire process of discovering and extracting a drop a finder module.
We evaluate the performance of our proposed method on images from the KEK and the MARCO data set.

\section{Methods}
\begin{figure}[htbp]
  \centering
  \includegraphics*[width=0.9\linewidth]{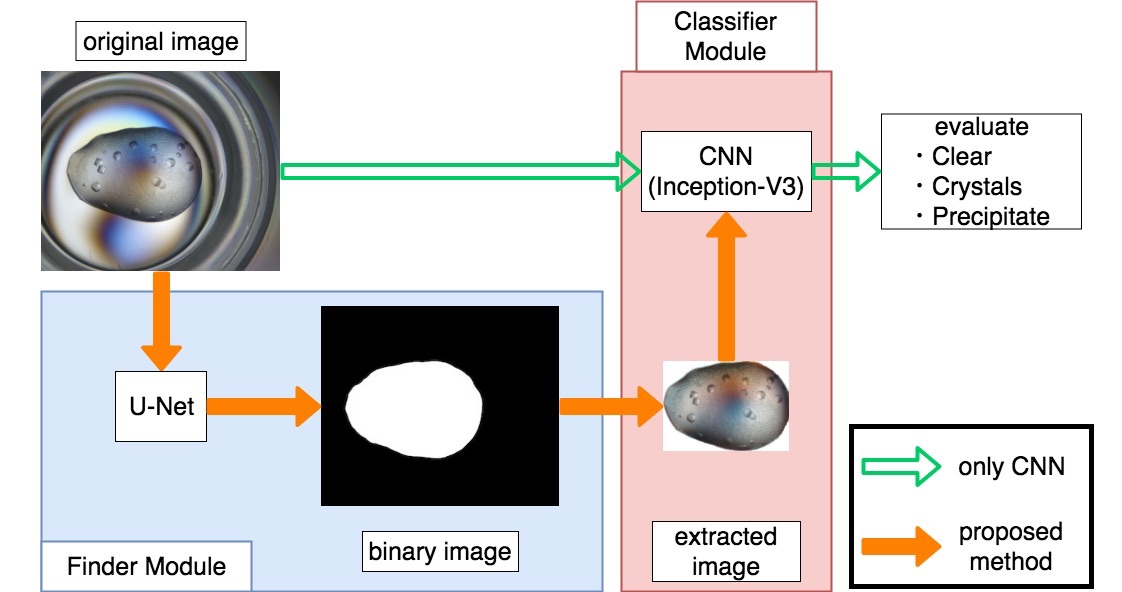}
  \caption{
  The overall of proposed method method.
  Top left image is the example of a target crystallization images.
  Top row is the process of using only CNN, and bottom row is that of the proposed method.
  Whereas blue region displays a finder module, and red region displays a classifier module.
  }
  \label{fig:proposed}
\end{figure}

In this section, the proposed method for automatic evaluation of crystallization images will be described.
An overview of the proposed method is shown in Figure \ref{fig:proposed}.
Our automatic recognition method consists of two parts: a finder module and a classifier module.
The finder module discovers the crystallization area and the classifier module classifies the state of crystallization.
In the finder module, images are cropped after capturing an area using U-Net.
The classifier module recognizes cropped images using CNN.

\subsection{Finder module}
\begin{figure}[htbp]
  \centering
  \includegraphics[width=0.8\linewidth]{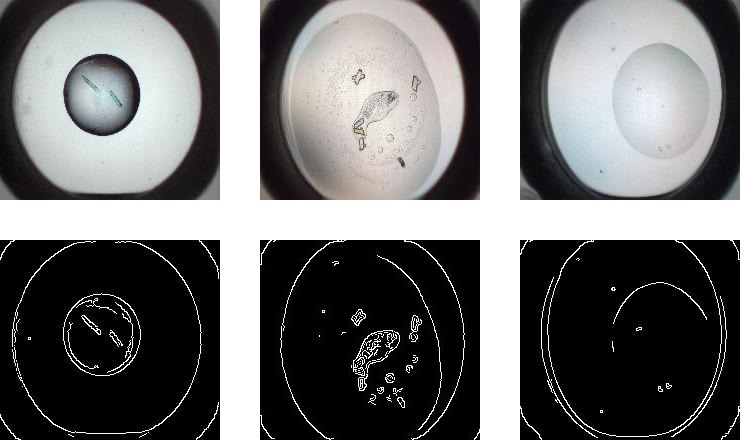}
  \caption{Examples of extracting a drop from each crystallization image using an edge filter with a same threshold.
    Original images (top).
    Edge images (bottom).
    The edge method successfully extracts the drop in the left image.
    The middle and right images illustrate failed attempts.
  }
  \label{fig:edge_result}
\end{figure}

It is difficult to discover the crystallization drop in an image correctly by using methods based on particular local features in the image.
The drop always exists in the crystallization image, however, its shape is complicated.
For example, we tried to apply the Canny edge filter \cite{Canny1986Edge} to find the crystallization drop (Figure \ref{fig:edge_result}).
The results of different images change severely depending on the threshold used.
On the other hand, when we use a model based CNN, the intricately shaped drop interfere with accuracy.
In addition, there are few examples of the more peculiar shapes, which makes it hard for the CNN to process them when dealing with full images.

In our method, we use U-Net \cite{Ronneberger2015UNet} which can perform image segmentation based on FCN with few images.
U-Net is divided into a contracting path and an expansive path.
The contracting path convolves the input with a filter like CNN and extracts the features necessary for segmentation from the image.
It repeats the downsampling unit $4$ times.
The unit consists of the repeated application of two $3 \times 3$ convolution layers which are followed by ReLU and a $2 \times 2$ max pooling layer with stride $2$.
Each time the unit is repeated, the number of feature channels doubles.
In the extracting path, upsampling the output is achieved by the contracting path and the process is repeated until the size of the output is equal to that of the input image.
Upsampling applies $2 \times 2$ convolution to the padded input and halves the number of feature channels.
The upsampled layer is concatenated with the contracting path's output, which corresponds to the unit.
After that, the extracting path applies two $3 \times 3$ convolutions, each followed by a ReLU.
This is repeated as many times as the contracting path.
In our finder module, we apply batch normalization to each convolution.

The Finder module extracts the recognized region after applying U-Net and sends the image to the classifier module.

\subsection{Classifier module}
The classifier module aims to classify the crystallization state from images.
While the primary focus of our research is to analyse the images from the KEK, in this work we aim to develop a method that can be employed more widely.
Therefore, we want to use a CNN model with a high precision on a variety of crystallization images.
With this in mind, we adopt the Inception-V3 CNN architecture for classifying crystallization images.

Inception-V3 \cite{Szegedy2016Inception} is characterized by complicated convolution called inception module, which distinguishes it from other CNN models.
In a conventional CNN, convolution is calculated using a filter of a certain size for input, and output to the next layer.
The Inception module convolves with multiple sized filters, combines the results obtained and outputs them to the next layer.
For example, VGG \cite{Simonyan2014VGG} convolved with a $3 \times 3$ filter, whereas in the original paper of the inception module, $1 \times 1$, $3 \times 3$, and $5 \times 5$ filters are combined and convolved.
Convolving the plurality of filters brings the enlargement of computation time.
To address this problem, applying $1 \times 1$ convolution before each convolution is calculated suppresses the computational complexity.
Bruno et al. \cite{Bruno2018Google} increased the size of the input by adding the convolution layer before Inception-V3 and demonstrated high discrimination accuracy.

Bruno et al. advocated that it is easy to grasp the various states of the crystallization by increasing the input size.
Our method can evaluate the crystallization states similarly.
The parameter set learned for an enormous data set called ImageNet \cite{Deng2009Imagenet} was used as the initial value to allow processing to be performed even with a small number of images.

\subsection{Parameters}
The proposed method was implemented using Keras.
Each of the modules were trained by Geforce GTX 108 0 Ti.
The finder module sets the batch size to $6$, the size of the learning image to $512 \times 512$, and the number of learning (epoch) to $300$.
For learning, a crystallization image and mask image, showing the drop portion in the image, were used as a set.
Adam was used as a method to optimize learning, and the learning rate was set to $1\mathrm{e}{-05}$.
The loss function was binary cross entropy and the mean IoU was used for the evaluation function.
IoU is an evaluation index as expressed by Equation (1), and the value obtained by evaluating (1) for each batch size and taking the class average value as the mean IoU.

\begin{equation}
  \textrm{IoU} = \frac{TP}{TP + FP + FN}
\end{equation}

where TP, FP, and FN stand for true positive, false positive, and false negative, respectively.

The classifier module has a batch size of $16$, a learning image size of $299 \times 299$, and an epoch of $300$.
Also RMSprop was used as an optimization method, and the learning rate was $1\mathrm{e}{-05}$.
The loss function was categorical cross entropy.

Data augmentation was applied for learning images.
In the finder module, gamma correction ($0.8 \leq \gamma \leq 1.2$), horizontal or vertical shift (shift with in $10\%$ of image size) and zoom ($0.9 \sim 1.1$ times) is applied.
In the classifier module, horizontal and vertical inversion and changing the color of the image by adding values for each channel of the image (pixel value $\pm 100$ range) are applied in addition to the horizontal and vertical shifts and zoom.

\section{Experimental Results}
In this section we demonstrate the accuracy of the proposed method, which combines the finder module and the classifier module.
We compared the performance of the CNN using the Full Image (Full Image CNN), using only the drop area which was found by manual analysis (Manual Finder CNN) and using the proposed finder module (Proposed).
We found good results with the MARCO data set and the KEK data set.
The MARCO data set was provided by the Collaborative Crystallization Center (C3), GlaxoSmithKline (GSK), Hauptman Woodward Medical Research Institute (HWI), Merck \& Co. (Merck), and Bristol-Myers Squibb (BMS).
This data set includes $493214$ images.
Since the purpose of this study was to achieve good accuracy with a small amount of images, we only use a small fraction of these images.
Additionally, the MARCO data set is divided into four states: Clear, Crystals, Precipitate, Others.
In order to guarantee fairly with the KEK data set, the images labeled as "others" were not included.

\subsection{Finder Module}
We verified the accuracy of extracted crystallzation regions by a finder module.
The finder module recognized and extracted the drop in the image, which was acquired from the MARCO or KEK data set.
The accuracy was evaluated by the mean IoU and the dice coefficient.
The dice coefficient indicates the similarity of sets.
For example, the dice coefficient between the set X and Y is expressed as follows:

\begin{equation}
  \textrm{dice} = \frac{2|X \cap Y|}{|X| + |Y|}
\end{equation}

\begin{figure}[htbp]
  \centering
  \includegraphics[width=0.9\linewidth]{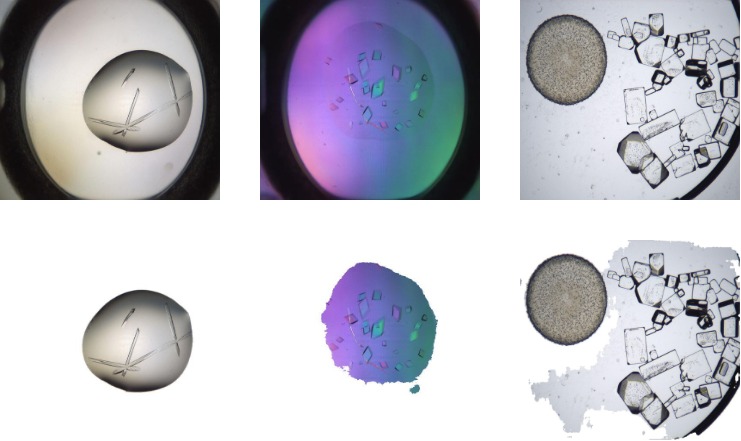}
  \caption{Results of finder module extracting the drop in the crystal image. The first row displays the original images; the second row displays the results of extracting with the finder module. The first column is an example of success, and the second and third columns illustrate failed attempts; the second column’s image is different from the depth of the drop, while the third column’s image is different rom the shape of the drop.The first and second column's drop used the Sitting Drop Vapor Diffusion Crystallization. The third column's used the sandwich method for protein crystallization.}
  \label{fig:failed_unet}
\end{figure}

We used $150 \sim 450$ images in the MARCO data set, and $190$ images in the KEK data set for training and evaluating training.
The number of images for evaluating training was $30\%$ of all images.
Table \ref{tbl:attent_result1} shows the finder module's evaluation values at each data set.
When calculating the mean IoU and the dice coefficient, we used $150$ images from MARCO and $88$ images from KEK where not used in model.
This result indicates that the finder module can find almost all the crystallization drops using only a few images for training.
However, the result of the MARCO is worse than the KEK.
The reasonable cause is insufficient data, but the most fundamental problem is the diversity in each of the organizations that compose the MARCO data set.
The finder module assumes images like the first column in Figure \ref{fig:failed_unet} and the KEK's image.
However, the MARCO's crystallization drop often appears unexpected shape for our method.
Therefore, U-Net failed extracting such as middle and right column in Figure \ref{fig:failed_unet}.

\begin{table}[htbp]
  \centering
    \begin{tabular}{|l|c|c|c|c|} \hline
      DataSet & Dice & Std(dice) & IoU & Std(IoU)\\ \hline
      MARCO $150$ & $0.922$ & $0.164$ & $0.884$ & $0.195$\\ \hline
      MARCO $300$ & $0.952$ & $0.092$ & $0.92$ & $0.133$\\ \hline
      MARCO $450$ & $0.942$ & $0.15$ & $0.915$ & $0.173$\\ \hline
      KEK & $0.974$ & $0.047$ & $0.953$ & $0.072$\\ \hline
    \end{tabular}
    \caption{Performance of the finder module. U-Net managed to find the drop in most images.}
    \label{tbl:attent_result1}
\end{table}

\begin{figure}[htbp]
  \centering
  \includegraphics[width=0.9\linewidth]{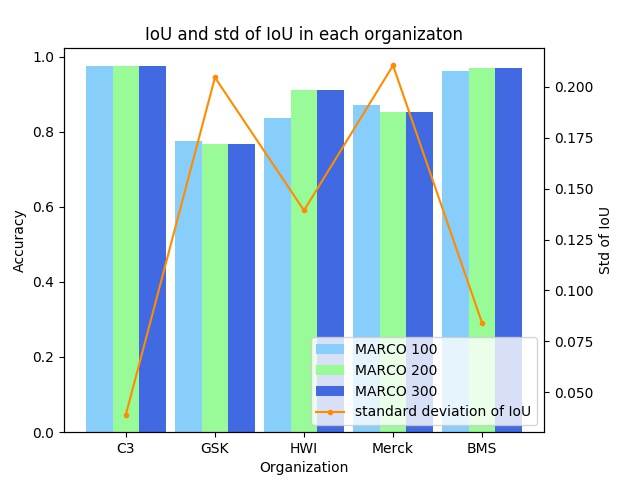}
  \caption{How the performance of the finder module changes according to the source organization of the data.
  The x-axis displays each orga¬nization, theleft  y-axis is the accuracy of U-Net, and the right y-axis is the standard deviation of the accuracy. Each orga¬nization provides crystal images taken by different equipment. Differences in facilities represent differences in crystal images.}
  \label{fig:attent_result2}
\end{figure}

\begin{figure*}[ht]
    \begin{minipage}{0.33\textwidth}
    \centering
    \includegraphics*[width=0.9\linewidth]{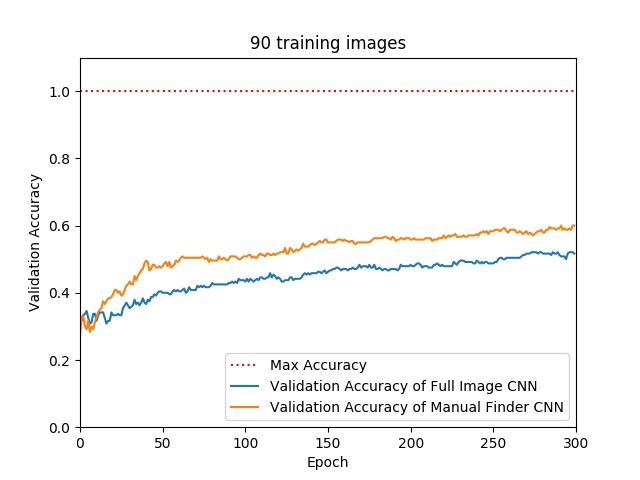}
    \includegraphics[width=0.9\linewidth]{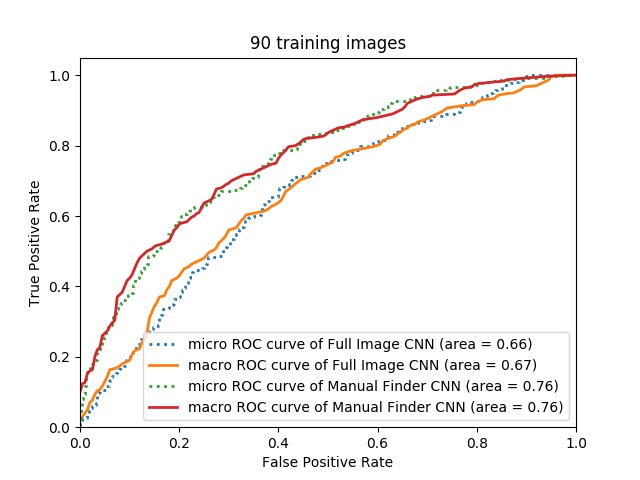}
    \end{minipage}
    \begin{minipage}{0.33\textwidth}
    \centering
    \includegraphics*[width=0.9\linewidth]{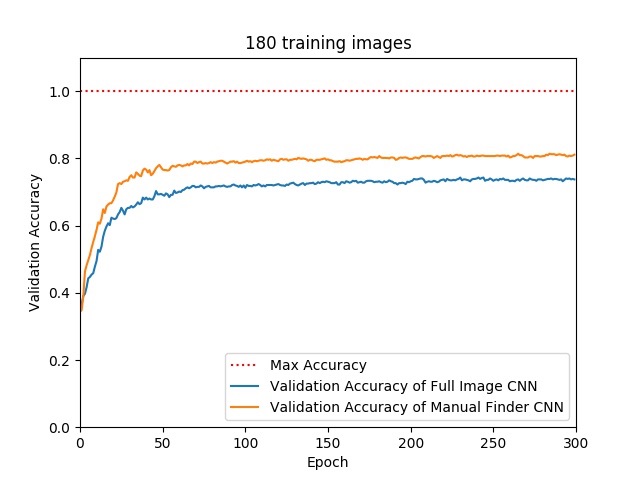}
    \includegraphics[width=0.9\linewidth]{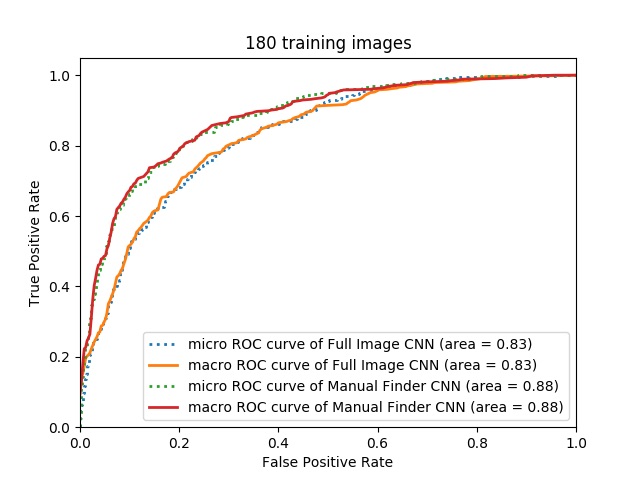}
    \end{minipage}
    \begin{minipage}{0.33\textwidth}
    \centering
    \includegraphics*[width=0.9\linewidth]{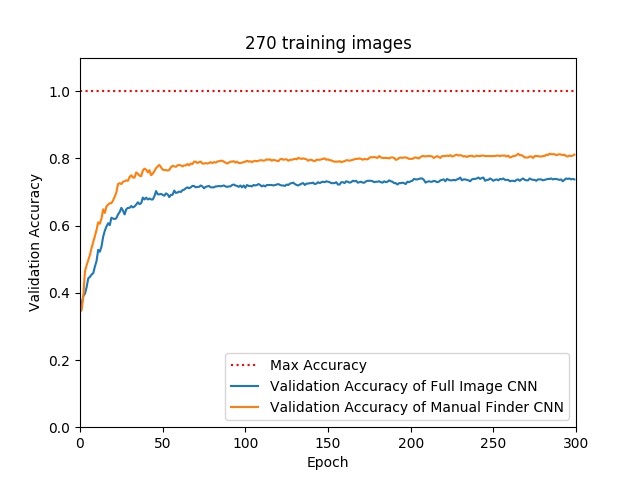}
    \includegraphics[width=0.9\linewidth]{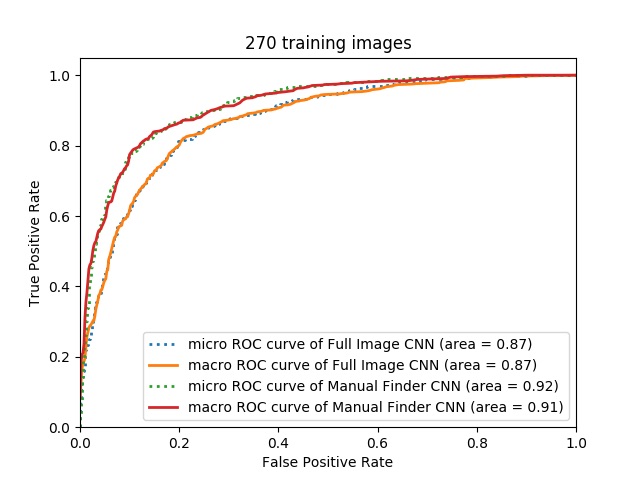}
    \end{minipage}
  \caption{
  Comparison between the Full Image CNN and using a manual finder module.
  The top row shows the accuracy of the validation data set through the training process and the bottom row shows ROC curves on the validation data.
  The size of the training and testing data sets is increased from left to right.
  We can observe that using the finder module improves both the validation accuracy and the area under the ROC curve.
  }
  \label{fig:recog_and_roc}
\end{figure*}

Figure \ref{fig:attent_result2} shows how the results change according to the organization that provided the images in the MARCO data set.
C3 and BMS have unified crystallization images, while GSK and Merck lack uniformity.
We see that images from uniform organization have higher accuracy and lower standard deviation.
To reduce this gap, it is necessary to make the gap uniform, or increase the number of heterogeneous data.

\subsection{Classifier module}
In this experiment, we show that the recognition performance of a CNN is improved by the presence of a finder module.
First, we verified how differences occurred in the accuracy when the crystallization region was completely extracted by the theoretical finder module.
We compared the accuracy of using only a CNN (Full Image CNN) and using a CNN after extracting the drop manually (Manual Finder CNN).
We used images which are used in experiments of the finder module.
When Full Image CNN and Manual Finder CNN trained for the images, the ratio of training and validation was $3$ to $2$.
Extracted drop images by Manual Finder CNN have only pixels based on the result of U-Net.
We measured the accuracy five times and calculate the mean value.
Figure \ref{fig:recog_and_roc} shows the result of this experiment.
We can see that the accuracy for the Manual Finder CNN is consistently greater than the Full Image CNN, and that area under the curve is about $5\%$ larger as well.
This indicates that using a finder module to discover the drop before giving the image to the classifier module is an effective approach.
However, manuallly extracting the drop from each image takes an enourmous amount of time, so we want to automate this process using the U-Net finder module.

\begin{figure}[htbp]
  \centering
  \includegraphics[width=0.9\linewidth]{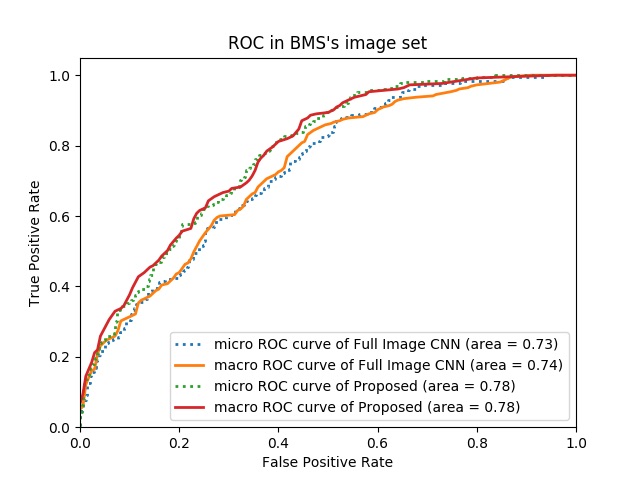}
  \caption{
  ROC curves on the BMS's data.
  Comparison between Full Image CNN and using U-Net finder module.}
  \label{fig:cut_result1}
\end{figure}

\begin{figure}[htbp]
  \centering
  \includegraphics[width=0.9\linewidth]{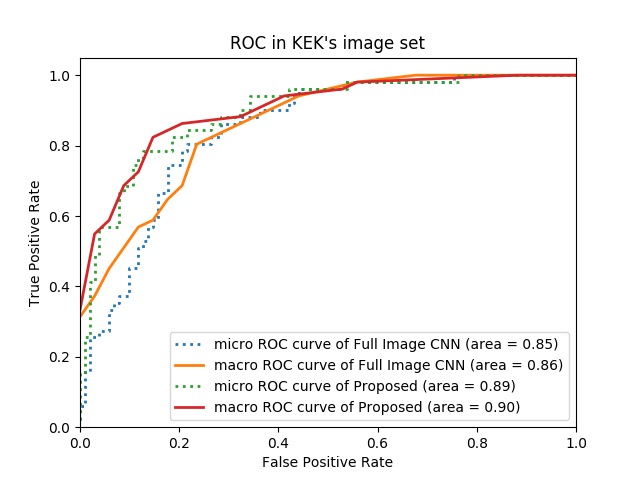}
  \caption{
  ROC curves on the KEK's data.
  Comparison between Full Image CNN and using U-Net finder module.}
  \label{fig:kek_result1}
\end{figure}

Next, we compared the performance of the U-Net finder module folowed by the classifier module (Proposed method) against the Full Image CNN.
For this experiment, we used only the BMS images from the MARCO data set (which showed the best finder result), using $99$ training images and $51$ validation images and all the images from the KEK data set being $90$ training images, $60$ validation images, and $88$ out-of-sample images used for calculating the final result.
The results of this experiments are in Figures \ref{fig:cut_result1} for the MARCO data set and \ref{fig:kek_result1} for the KEK data set.
We observed that using the finder module increases the AUC by about $5\%$.

\begin{figure}[htbp]
  \centering
  \includegraphics[width=0.6\linewidth]{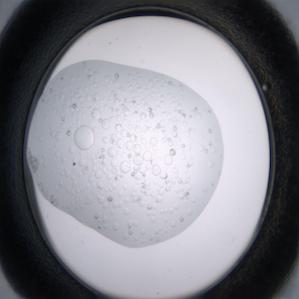}
  \includegraphics[width=0.6\linewidth]{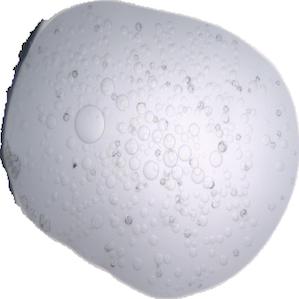}
  \caption{
  Examples of the good result which Proposed classified correctly.
  The top image is the whole image which is used Full Image CNN.
  The bottom image is extracted from the top image by the finder module.
  }
  \label{fig:good_example}
\end{figure}

Figure \ref{fig:good_example} shows the image which Proposed classified the correct state.
The image's state is Crystals, however, it is difficult to tell Precipitate from Crystals by observing the whole image.
Whereas Full Image CNN classified as Precipitate, the enlargement drop helped to classify as Crystals in Proposed.

\begin{figure}[htbp]
  \centering
  \includegraphics[width=0.9\linewidth]{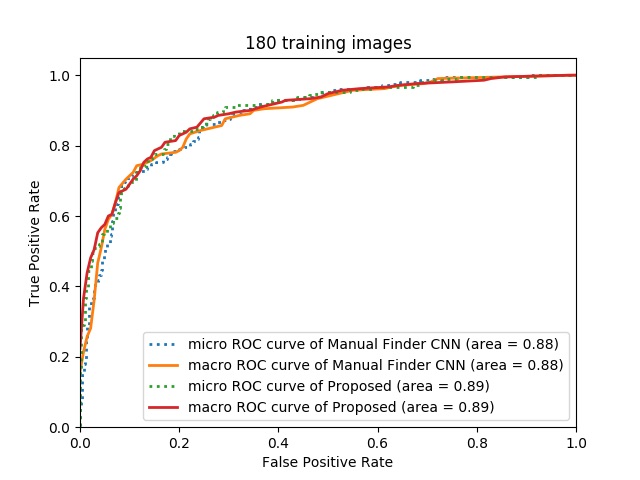}
  \caption{
  Comparison using between manual and U-Net finder module.}
  \label{fig:compare_result1}
\end{figure}

Finally, we compared the results between the manual finder module and the U-Net finder module.
Figure \ref{fig:compare_result1} shows that both methods show an equivalent performance.
This indicates that we can use the automatic finder module to avoid the costly process of extracting each drop area manually.
These three experiments demonstrate the efficacy of the proposed method.

\section{Conclusion}
In this paper, we investigate the idea of using a finder module that discovers the key region of an image before using a CNN on a classification task.
Our target application is the classification of crystallization images.
We first showed that extracting the key region (drop) manually increased the accuracy of the CNN.
However, as the manual extraction of the target region is a time intensive task, we proposed an automatic finder module using the U-Net.
Therefore, our proposed method for accurate classification of crystallization images uses U-Net as the finder module and Inception-V3 as the classifier module.
We showed that this proposed method achieves about $5\%$ increase of the AUC for BMS data from the MARCO data set and the KEK data set.
At the moment we are investigating other models which could be used to improve the performance of each module and we want to also increase the scale of our experiments on more diverse data sets.

\bibliography{refmiura}
\bibliographystyle{aaai}
\end{document}